\documentclass[conference]{IEEEtran}
\usepackage{amsmath,amssymb,amsfonts}
\usepackage{algorithmic}
\usepackage{placeins}
\usepackage{booktabs}
\usepackage{adjustbox}
\usepackage{array, multicol, multirow}
\usepackage{tabularray}
\usepackage[caption=false,font=footnotesize,labelfont=sf,textfont=sf]{subfig}
\usepackage{textcomp}
\usepackage{stfloats}
\usepackage{url}
\usepackage{verbatim}
\usepackage{graphicx}
\usepackage{cite}
\usepackage[acronym]{glossaries}
\usepackage{xcolor}

\newacronym{TMS}{TMS}{Transcranial Magnetic Stimulation}
\newacronym{Robo-TMS}{Robo-TMS}{Robot-assisted Transcranial Magnetic Stimulation}
\newacronym{MRI}{MRI}{Magnetic Resonance Imaging}
\newacronym{CAD}{CAD}{Computer-Aided Design}
\newacronym{ROS}{ROS}{Robot Operating System}
\newacronym{GUI}{GUI}{Graphical User Interface}
\newacronym{MRML}{MRML}{Medical Reality Modelling Language}
\newacronym{VTK}{VTK}{Visualization Toolkit}
\newacronym{URDF}{URDF}{Unified Robot Description Format}
\newacronym{STL}{STL}{stereolithography}
\newacronym{XML}{XML}{Extensible Markup Language}

\begin{document}
\bstctlcite{IEEEtran:BSTcontrol}

%%%%%%%%%%%%%%%%%%%%  Title  %%%%%%%%%%%%%%%%%%%%
\title{
SlicerRoboTMS: An Open-Source 3D Slicer Extension for Robot-Assisted Transcranial Magnetic Stimulation \\
% \thanks{For the purpose of open access, the author has applied a Creative Commons Attribution (CC BY) license to any Author Accepted Manuscript version arising.}
}

\author{
\IEEEauthorblockN{1\textsuperscript{st} Wenzhi Bai}
\IEEEauthorblockA{\textit{Department of Electrical and Electronic Engineering} \\
\textit{University of Manchester}\\
Manchester, U.K. \\
wenzhi.bai@manchester.ac.uk}
\and
\IEEEauthorblockN{2\textsuperscript{nd} Yituo Guo}
\IEEEauthorblockA{\textit{Department of Electrical and Electronic Engineering} \\
\textit{University of Manchester}\\
Manchester, U.K. \\
yituo.guo@postgrad.manchester.ac.uk}

\and
\IEEEauthorblockN{3\textsuperscript{rd} Bhaskar Basu}
\IEEEauthorblockA{\textit{Rehabilitation Medicine}\\
\textit{Manchester University NHS Foundation Trust} \\
Manchester, U.K. \\
bhaskar.basu@mft.nhs.uk}
\and
\IEEEauthorblockN{4\textsuperscript{th} Andrew Weightman}
\IEEEauthorblockA{\textit{Department of Mechanical and Aerospace Engineering} \\
\textit{University of Manchester}\\
Manchester, U.K. \\
andrew.weightman@manchester.ac.uk}
\and
\IEEEauthorblockN{5\textsuperscript{th} Zhenhong Li}
\IEEEauthorblockA{\textit{Department of Electrical and Electronic Engineering} \\
\textit{University of Manchester}\\
Manchester, U.K. \\
zhenhong.li@manchester.ac.uk}
}

\maketitle

% \begingroup
% \let\thefootnote\relax
% \footnotetext{© 2026 IEEE. Personal use of this material is permitted. Permission from IEEE must be obtained for all other uses, in any current or future media, including reprinting/republishing this material for advertising or promotional purposes, creating new collective works, for resale or redistribution to servers or lists, or reuse of any copyrighted component of this work in other works. For the purpose of open access, the author has applied a Creative Commons Attribution (CC BY) license to any Author Accepted Manuscript version arising.}
% \endgroup

%%%%%%%%%%%%%%%%%%%%  Abstract  %%%%%%%%%%%%%%%%%%%%
\begin{abstract}
\Gls{Robo-TMS} is an image-guided robotic intervention that enhances the accuracy and reproducibility of conventional \gls{TMS}, a widely used non-invasive brain stimulation procedure in clinical treatment and neuroscience research. Despite its potential, the development of \gls{Robo-TMS} remains challenging due to the need for multidisciplinary expertise spanning medical imaging, computer vision, and robotics. This paper presents \textit{SlicerRoboTMS}, an open-source 3D Slicer extension that provides a unified interaction infrastructure for \gls{Robo-TMS} research. By leveraging 3D Slicer's medical image computing and visualisation capabilities, the extension supports \gls{MRI}-based neuronavigation and interfaces with robotic systems through standardised communication protocols and configurable system descriptions. An example integration is presented to demonstrate how SlicerRoboTMS can be incorporated into a representative \gls{Robo-TMS} workflow. Designed to support diverse hardware configurations and rapid prototyping, SlicerRoboTMS lowers the barrier to entry and facilitates reproducible and extensible research in \gls{Robo-TMS}. The extension is available at \url{https://github.com/OpenRoboTMS/SlicerRoboTMS}.
\end{abstract}


%%%%%%%%%%%%%%%%%%%%  Introduction  %%%%%%%%%%%%%%%%%%%%
\section{Introduction}
\acrfull{TMS} is a non-invasive neuromodulation procedure that uses rapidly changing magnetic fields to induce electric currents in targeted brain regions~\cite{edwards2024practical}. It is clinically validated for the treatment of neuropsychiatric disorders such as major depressive disorder and obsessive-compulsive disorder, and has demonstrated growing potential in neurorehabilitation for conditions including stroke and Parkinson's disease~\cite{vucic2023clinical, cohen2022visual, rossi2021safety}. In addition, \gls{TMS} is widely used for functional brain mapping in both clinical and research contexts, including pre-surgical planning for brain tumour resection~\cite{sollmann2021mapping, haddad2020preoperative} and neuroscientific investigations of cortical organisation~\cite{kahl2023reliability, giuffre2021reliability}. To improve targeting accuracy and procedural consistency, \acrfull{Robo-TMS} integrates neuronavigation with a robotic manipulator to position the stimulation coil relative to subject-specific anatomy~\cite{fichtinger2022imageguided}. By compensating for operator variability and subject head motion, \gls{Robo-TMS} can enhance therapeutic reliability and procedural efficiency compared to manual \gls{TMS}~\cite{bai2025robotassisted}. As a result, \gls{Robo-TMS} has emerged as a rapidly growing research area with the potential to transform conventional \gls{TMS} workflows.

Despite this promise, the development of \gls{Robo-TMS} systems remains challenging. Effective \gls{Robo-TMS} requires tight integration of medical image data, spatial tracking, and robotic control. However, \acrfull{MRI}-based neuronavigation, which is essential for target identification, is difficult to incorporate into robotic development environments such as the \gls{ROS}~\cite{macenski2022robot}. While \gls{ROS} provides widely used software libraries and tools for robotics, it does not natively support image-guided clinical workflows. This disconnect increases system complexity and raises the barrier to entry for system development. 3D Slicer is a widely adopted open-source platform for medical image computing and image-guided therapy research~\cite{fedorov20123d}. It provides comprehensive tools for medical image processing and visualisation, together with extensible interfaces for integrating external hardware and software systems. In addition, several middleware frameworks have been proposed to facilitate the integration of medical imaging workflows with \gls{ROS}~\cite{connolly2024slicerros2, frank2017rosigtlbridge, tokuda2009openigtlink}. These capabilities have supported a wide range of image-guided interventions~\cite{sahu2024entri, moreira2021vivo, choueib2019evaluation, ungi2016navigated}, and recent work has demonstrated feasibility for \gls{TMS}-related research through extensions such as SlicerTMS~\cite{franke2024slicertms}. However, to the authors' knowledge, no open-source \gls{Robo-TMS} extension currently exists within the 3D Slicer ecosystem, which limits access to advanced medical image computing capabilities and hinders broader participation.

Existing \gls{Robo-TMS} systems~\cite{yang2025neuromodulation, kebria2023hapticallyenabled, noccaro2021development, ginhoux2013custom, pennimpede2013hot} are often device-specific and rely on proprietary software, further fragmenting research efforts. To address this gap, we present \textit{SlicerRoboTMS}, an open-source 3D Slicer extension that provides a unified interaction infrastructure for \gls{Robo-TMS} research. The proposed extension supports the integration of \gls{MRI}-based neuronavigation with robotic systems through standardised interfaces, reducing system re-engineering and facilitating extensible research across diverse \gls{Robo-TMS} applications.

The contributions of this paper are as follows:
\begin{itemize}
    \item the first open-source 3D Slicer extension designed specifically for \gls{Robo-TMS} research;
    \item a modular and configurable extension that supports extensible \gls{Robo-TMS} applications;
    \item an example integration illustrating the practical use of the proposed extension.
\end{itemize}

The proposed extension supports open, extensible, and rapid prototyping of \gls{Robo-TMS} systems, facilitating reproducible and collaborative research in this emerging field.

%%%%%%%%%%%%%%%%%%%%  Extension Overview  %%%%%%%%%%%%%%%%%%%%
\section{Extension Overview}
This section provides an overview of the SlicerRoboTMS extension within a \gls{Robo-TMS} system, focusing on its design considerations, integration context, and user interaction.

\subsection{Design Considerations}
\gls{Robo-TMS} requires tight integration of \gls{MRI} data, spatial tracking, and robotic control within a unified software environment. SlicerRoboTMS is designed with the following considerations to support flexible, extensible, and reproducible research.

\subsubsection{Openness and extensibility}
SlicerRoboTMS provides fully open-source code together with example integration implementations, enabling transparent inspection, modification, and extension of the software. This supports rapid prototyping, reproducible research, and community-driven development.

\subsubsection{Modularity and compatibility}
The software follows the modular design principles of 3D Slicer to ensure compatibility with its existing ecosystem. By leveraging established 3D Slicer data structures and communication protocols, the extension integrates with existing tools and supports seamless incorporation into broader research workflows.

\subsubsection{Hardware agnosticism}
Standardised communication interfaces and configuration-based hardware descriptions are used to decouple image-guided workflows from hardware-specific implementations, allowing the extension to interface with a wide range of external devices.

\subsubsection{Real-time performance}
SlicerRoboTMS employs configurable visualisation update rates and event-driven data exchange to support real-time updates of robotic poses, target locations, and anatomical context during system operation.

\subsection{Integration Architecture Overview}
\begin{figure}[!htb]
    \centering
    \includegraphics[width=1\linewidth]{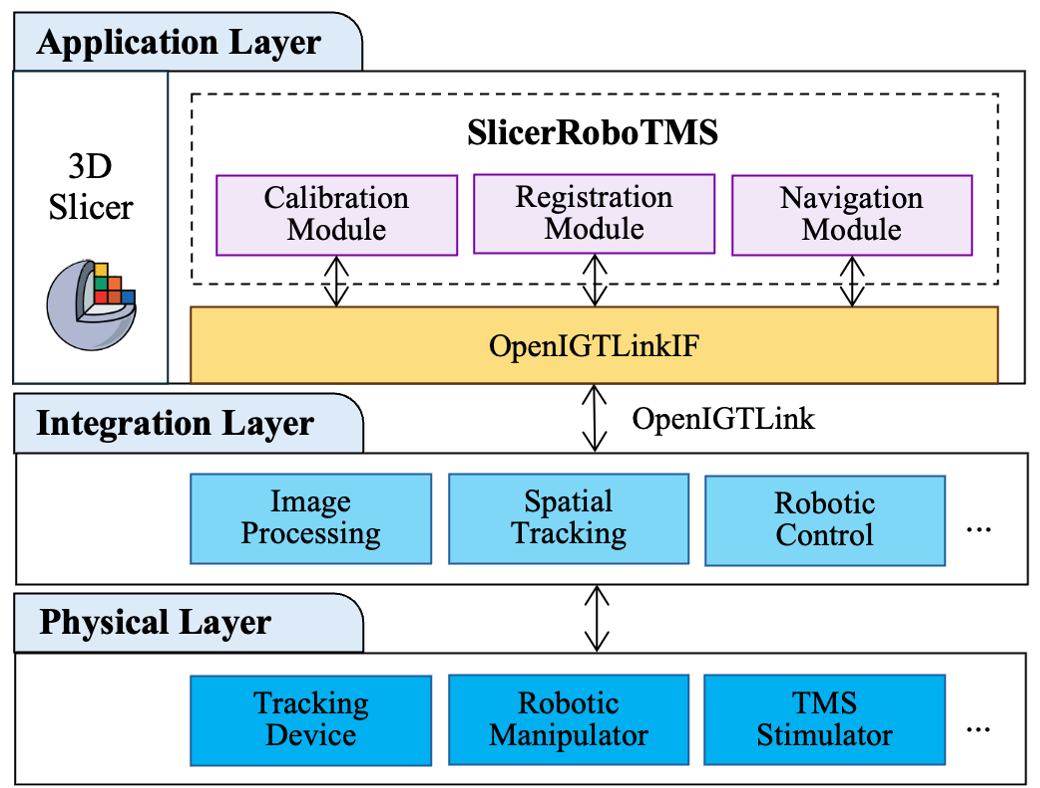}
    \caption{Contextual system architecture illustrating the role of SlicerRoboTMS within a generic \gls{Robo-TMS} setup. The architecture adopts a layered design comprising an application layer based on 3D Slicer, an integration layer responsible for system coordination, and a physical layer including tracking devices, robotic manipulators, and \gls{TMS} stimulators. SlicerRoboTMS operates as an extension within the application layer and provides three functional modules. Standardised communication interfaces enable real-time data exchange between layers; in particular, OpenIGTLink is used for communication between the application and integration layers via the OpenIGTLink Interface (OpenIGTLinkIF) module in 3D Slicer.}
    \label{fig:system_architecture}
\end{figure}

SlicerRoboTMS is organised as an extension comprising three functional modules: a calibration module, a registration module, and a navigation module. These modules operate independently within 3D Slicer while collectively supporting a typical \gls{Robo-TMS} workflow. To place the extension in context, Fig.~\ref{fig:system_architecture} illustrates the layered system architecture in which SlicerRoboTMS operates, separating medical image-based interaction, system coordination, and physical device execution.

Within this architecture, SlicerRoboTMS operates exclusively in the application layer and provides the primary interface for system monitoring, \gls{MRI}-based neuronavigation, and user interaction. The application layer is built on 3D Slicer, which offers robust support for medical image visualisation, spatial transforms, and interactive planning, making it well suited for neuronavigation-driven \gls{Robo-TMS} workflows. Implementing SlicerRoboTMS as a native 3D Slicer extension allows the reuse of established data structures and visualisation capabilities, while enabling communication with external components through standardised interfaces.

The integration and physical layers are included to contextualise how SlicerRoboTMS interfaces with external system components. The integration layer coordinates image processing, spatial tracking, and robotic control, and communicates with the application layer using OpenIGTLink~\cite{tokuda2009openigtlink} to maintain loose coupling between medical image computing and device-specific control. The physical layer comprises tracking devices, robotic manipulators, and \gls{TMS} stimulators, which are accessed indirectly through the integration layer. This layered architecture supports flexible and extensible \gls{Robo-TMS} research workflows while allowing SlicerRoboTMS to remain independent of specific hardware configurations.

\subsection{User Interaction and Visualisation}

\begin{figure*}[!htb]
    \centering
    \includegraphics[width=1\linewidth]{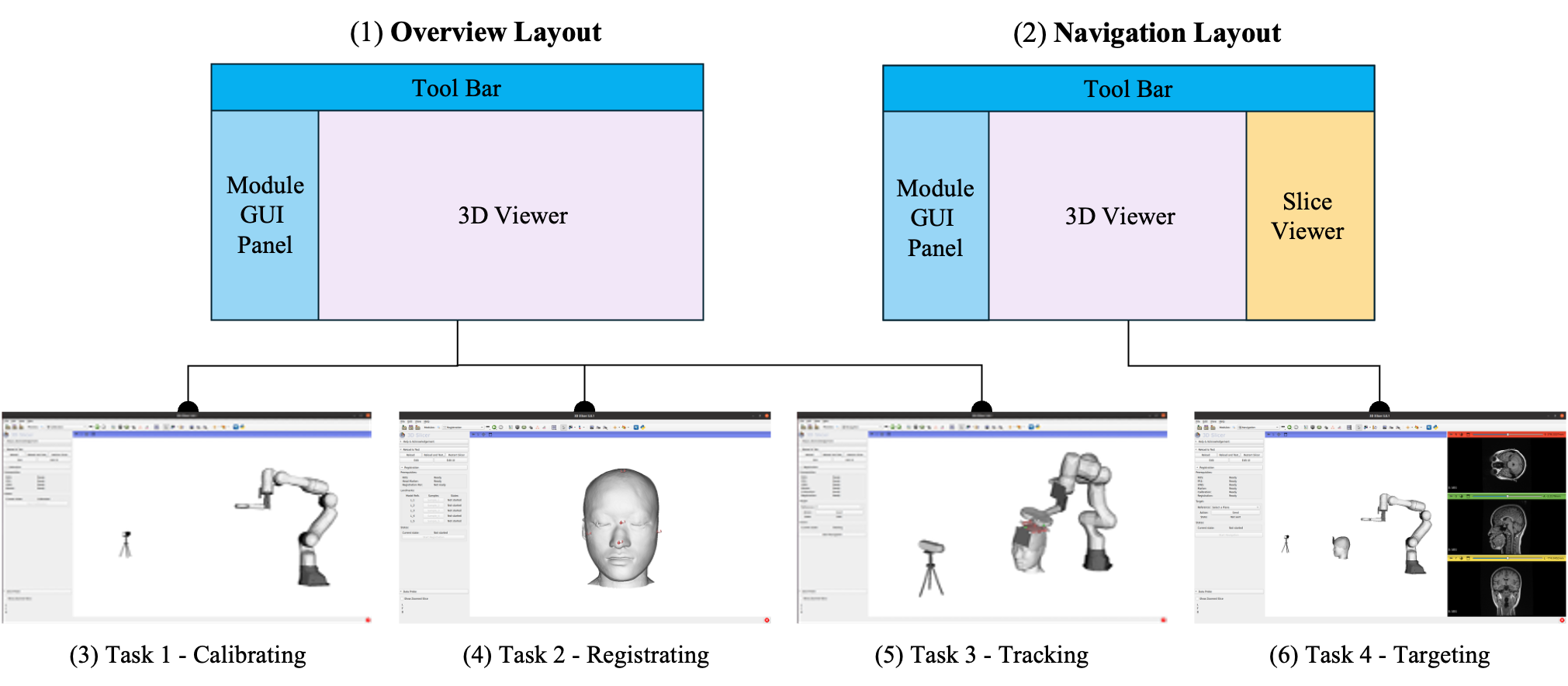}
    \caption{User interaction and visualisation layouts provided by SlicerRoboTMS. Two primary layouts are supported: (1) a system overview layout with a 3D view for global monitoring, and (2) a navigation layout combining a 3D view with orthogonal slice views for \gls{MRI}-guided neuronavigation. Task-specific views corresponding to different modules and workflow stages are illustrated in (3)-(6). }
    \label{fig:gui_layouts}
\end{figure*}

Effective \gls{Robo-TMS} interaction requires both real-time feedback on system state and accurate \gls{MRI}-based neuronavigation. SlicerRoboTMS addresses these requirements through a layout-based \gls{GUI} design that presents task-relevant information while maintaining a clear and consistent user experience, as illustrated in Fig.~\ref{fig:gui_layouts}.

The \gls{GUI} is implemented within the 3D Slicer environment and builds directly on its native view layout framework, which supports synchronised 3D visualisation and orthogonal medical image slices. SlicerRoboTMS customises these existing layouts to support different workflow tasks. A system overview layout, derived from the standard 3D-only layout in 3D Slicer, supports global monitoring by visualising the robotic manipulator, stimulation coil, tracking data, and anatomical models. A navigation layout, adapted from the conventional widescreen layout, combines a 3D view with orthogonal \gls{MRI} slice views to enable precise target localisation relative to subject-specific anatomy.

During calibration, registration, and navigation, the interface selects the layout most appropriate for the task and provides tailored module panels to present essential parameters and system states. By leveraging and extending native 3D Slicer visualisation capabilities, SlicerRoboTMS enables intuitive user interaction while remaining consistent with established \gls{Robo-TMS} workflows.

%%%%%%%%%%%%%%%%%%%%  Extension Implementation  %%%%%%%%%%%%%%%%%%%%
\section{Extension Implementation}
This section details the implementation of the SlicerRoboTMS extension, focusing on data management and interface mechanisms for integration with external components.

\subsection{Communication Protocol}
SlicerRoboTMS serves as the interaction hub for image-guided \gls{Robo-TMS} workflows, exchanging spatial information, system states, and user-defined commands with the integration layer. To enable real-time data exchange while maintaining loose coupling between software components, OpenIGTLink is adopted as the communication protocol between the application and integration layers. OpenIGTLink is a TCP/IP-based protocol designed for real-time data exchange in image-guided medical applications and is supported in 3D Slicer via the OpenIGTLink Interface (OpenIGTLinkIF) module. This enables SlicerRoboTMS to interface with tracking and robotic systems without embedding hardware-specific dependencies.

All data in 3D Slicer are managed using the \gls{MRML}~\cite{fedorov20123d}, where each data entity is represented as a node within a scene and updated through event-driven mechanisms. In SlicerRoboTMS, \gls{MRML} nodes are mapped to standard OpenIGTLink message types to support different aspects of the \gls{Robo-TMS} workflow:
\begin{itemize}
    \item \texttt{TRANSFORM} messages transmit real-time spatial information, such as tracking data and robot poses, enabling dynamic visualisation;
    \item \texttt{POLYDATA} messages represent geometrical objects, such as registration fiducials;
    \item \texttt{STATUS} messages convey system states and execution commands for monitoring and user interaction.
\end{itemize}

Each message is assigned a unique identifier to ensure consistent handling within the communication stream. This design preserves a clear separation between image-guided interaction in the extension and hardware-specific control logic implemented in the integration layer, while supporting flexible and extensible system integration.

\subsection{Transform Management}
Accurate and consistent spatial relationships are essential for \gls{Robo-TMS}, as dynamic visualisation depends on continuous updates of tracking and robotic pose information. In the overall workflow, real-time spatial data are maintained in the integration layer and transmitted to the application layer via OpenIGTLink, where visualisation is driven by transform updates within the 3D Slicer scene. SlicerRoboTMS leverages 3D Slicer's native transform hierarchy to manage these spatial relationships in a manner consistent with image-guided workflows.

Within SlicerRoboTMS, all visualised entities are represented in the \gls{MRML} scene using separate model and transform nodes to define geometry and relative pose, respectively. Model geometry is specified using \gls{STL} files, while corresponding transform nodes encode spatial relationships between coordinate frames using linear transformation matrices. These transforms are organised into a hierarchical tree structure, allowing updates at any level to propagate automatically and preserve spatial consistency. The default transform hierarchy is initialised from a \gls{URDF} configuration file, which specifies model paths, parent-child relationships, frame identifiers, and visual properties. During operation, updated transforms received via OpenIGTLink modify the associated transform nodes and trigger scene update events, enabling real-time and coherent visualisation of the robot, stimulation coil, and tracked anatomy.

\subsection{Scene Integration}
In SlicerRoboTMS, the \gls{MRML} scene provided by 3D Slicer is used as a unified data container to manage imaging data, models, spatial relationships, and interaction states within the extension. The scene includes \gls{MRI} volumes, model nodes representing physical components, and transform nodes defining spatial relationships among these components. Together, these elements form a coherent virtual representation of the \gls{Robo-TMS} setup, enabling consistent visualisation of anatomy, tracking data, and robotic pose within a single coordinate framework. Each functional module in SlicerRoboTMS initialises its scene from predefined configurations, while real-time states are maintained in the integration layer and synchronised with the extension during operation. Model geometries and transform hierarchies are initialised from \gls{URDF} configuration files in \gls{XML} format, and system-level parameters are specified using Python configuration files.

Scene updates are performed through a periodic synchronisation mechanism driven by a Qt timer, which triggers updates at a configurable rate of 30~Hz. In the present setup, this update rate supports real-time visualisation and is comparable to the refresh frequency of commonly used optical tracking systems. Incoming spatial updates modify the corresponding transform nodes in the scene and trigger event-driven updates to ensure responsive visual feedback. All models and transforms are defined and loaded in the RAS (Right, Anterior, Superior) coordinate system to align with 3D Slicer's spatial conventions and common robot development environments, thereby reducing coordinate ambiguity and simplifying integration.

%%%%%%%%%%%%%%%%%%%%  Example Integration  %%%%%%%%%%%%%%%%%%%%
\section{Example Integration}
This section presents an example integration to demonstrate how SlicerRoboTMS can be integrated into an experimental \gls{Robo-TMS} system.

\subsection{Experimental Setup}
\begin{figure*}[!htb]
    \centering
    \includegraphics[width=0.9\linewidth]{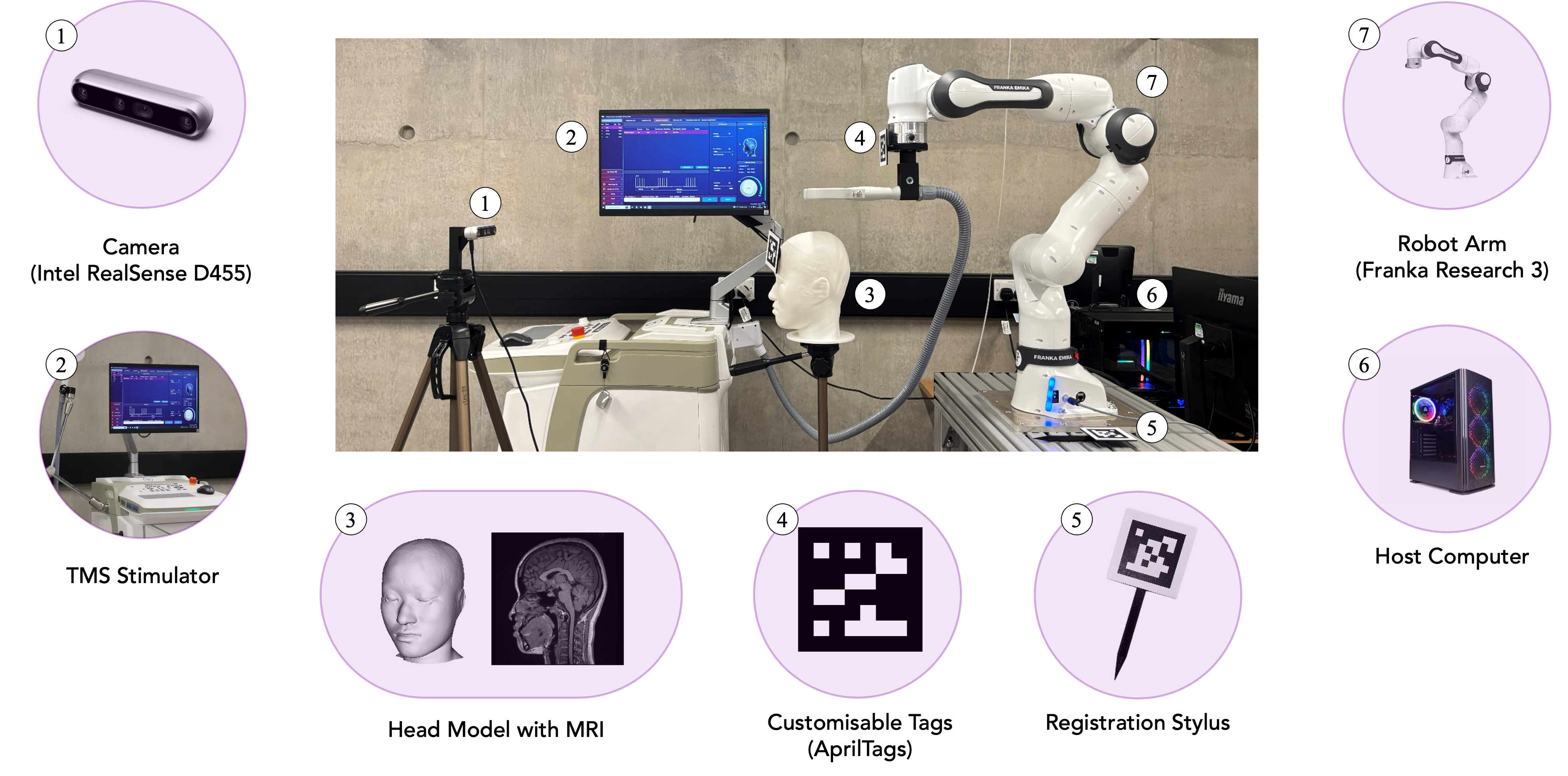}
    \caption{Experimental setup used for the example integration. The configuration includes: (1) an optical tracking camera (Intel RealSense D455), (2) a \gls{TMS} stimulator, (3) a 3D-printed head phantom generated from subject-specific \gls{MRI} data, (4) customisable fiducial tags (AprilTags), (5) a 3D-printed registration stylus with the fiducial tag, (6) a host computer, and (7) a robotic manipulator (Franka Research 3). Numbered annotations indicate the main components, which are shown in separate subfigures for clarity.}
    \label{fig:experimental_setup}
\end{figure*}

An example \gls{Robo-TMS} setup, shown in Fig.~\ref{fig:experimental_setup}, is implemented to illustrate the practical use of SlicerRoboTMS.

Spatial tracking is performed using an Intel RealSense D455 camera and customisable fiducial tags (AprilTags~\cite{krogius2019flexible}) attached to the robotic end-effector, head phantom, and a registration stylus. The colour camera operates at a resolution of $1280 \times 800$ pixels and a frame rate of 30~Hz, enabling real-time pose estimation. A Franka Research 3 robotic manipulator~\cite{haddadin2024franka} positions the \gls{TMS} coil relative to the 3D-printed head phantom. The setup runs on a host computer equipped with an Intel Core i9 processor and 64~GB of memory, using Ubuntu~20.04 with real-time kernel patches and \gls{ROS} Noetic. This configuration represents a typical laboratory environment and is used to demonstrate the integration and operation of SlicerRoboTMS.

\subsection{Integration Details}
The integration architecture of the example setup is illustrated in Fig.~\ref{fig:example_architecture}.

\begin{figure}[!htb]
    \centering
    \includegraphics[width=1\linewidth]{Pictures/Example_Architecture.png}
    \caption{Integration architecture of the example SlicerRoboTMS-based \gls{Robo-TMS} setup. The integration layer, implemented using \gls{ROS}, interfaces with physical devices and maintains real-time system states through distributed nodes. Bidirectional communication between the integration and application layers is achieved via the ROS-IGTL-Bridge (server) and OpenIGTLinkIF (client) using the OpenIGTLink protocol, enabling synchronised visualisation, navigation, and robotic control within 3D Slicer.}
    \label{fig:example_architecture}
\end{figure}

In the example integration, system configurations and hardware descriptions are defined using configuration files to initialise SlicerRoboTMS with the appropriate models, transforms, and parameters. The integration layer is implemented using \gls{ROS}, which provides a modular framework based on distributed processes, referred to as nodes. \gls{ROS} topics are used for real-time data streaming, and the \texttt{tf} library is employed to manage spatial transforms between coordinate frames. Communication between the \gls{ROS}-based integration layer and SlicerRoboTMS is achieved using the ROS-IGTL-Bridge~\cite{frank2017rosigtlbridge}, which translates \gls{ROS} messages into OpenIGTLink messages transmitted over a TCP/IP connection to 3D Slicer. Conversely, messages generated by the extension are converted into \gls{ROS} messages and published within the \gls{ROS} network. This bidirectional translation enables the exchange of spatial data, system states, and execution commands while maintaining loose coupling between image-guided interaction and robotic control.

Within the integration layer, five independent \gls{ROS} nodes are used to process and maintain system data. A detector node estimates fiducial poses from camera images. A calibrator node computes spatial relationships among different components. A register node establishes the transform between \gls{MRI} image space and the physical head phantom. A tracker node converts stimulation targets defined in SlicerRoboTMS into corresponding robotic end-effector poses. Finally, a controller node executes robotic control based on the desired poses. Together, these nodes form a modular and extensible integration layer that enables the practical use of SlicerRoboTMS in a \gls{Robo-TMS} workflow.

\subsection{Observations}
The example integration indicates that incorporating SlicerRoboTMS into a \gls{Robo-TMS} setup primarily involves development within the integration layer. This layer encapsulates system-specific algorithms, including tracking, calibration, registration, and robotic control, and interfaces directly with physical devices. Consequently, algorithm development and hardware control can be carried out independently of medical image visualisation and user interaction, which are handled by SlicerRoboTMS within 3D Slicer.

Based on the example integration, three aspects are essential for effective use of SlicerRoboTMS:
\begin{itemize}
    \item Communication between the integration layer and the extension should rely on the OpenIGTLink protocol as the sole interface for exchanging spatial transforms, target definitions, and system states.
    \item Using \gls{ROS} as the foundation of the integration layer is a practical choice due to its mature robotic development ecosystem and native support for distributed processing and transform management. Tools such as the ROS-IGTL-Bridge facilitate bidirectional message mapping between \gls{ROS} and 3D Slicer.
    \item System configuration should be aligned with the physical setup. Configuration files define model meshes, \gls{MRI} volumes, and transform hierarchies to ensure correct scene initialisation and consistent spatial interpretation.
\end{itemize}

By providing ready-made modules for medical image visualisation and user interaction, SlicerRoboTMS enables researchers to avoid re-engineering the \gls{Robo-TMS} interface and instead focus on developing and validating calibration, registration, and robotic control algorithms within a unified research framework.

%%%%%%%%%%%%%%%%%%%%  Conclusion  %%%%%%%%%%%%%%%%%%%%
\section{Conclusion}
This paper presents SlicerRoboTMS, an open-source 3D Slicer extension designed to support medical image–guided \gls{Robo-TMS} research. The extension provides a unified interaction infrastructure for \gls{Robo-TMS}, enabling \gls{MRI}-based neuronavigation, real-time visualisation, and user interaction within the 3D Slicer ecosystem. The example integration demonstrates that researchers can implement specific calibration, registration, and robotic control algorithms externally while relying on SlicerRoboTMS for consistent visualisation and interaction, helping reduce development effort. By lowering the barrier to developing \gls{Robo-TMS} systems, SlicerRoboTMS allows researchers to focus on methodological innovation rather than software infrastructure. The extension provides a practical foundation for reproducible research and supports future advances in \gls{Robo-TMS}.

% \section*{Acknowledgment}

% The preferred spelling of the word ``acknowledgment'' in America is without 
% an ``e'' after the ``g''. Avoid the stilted expression ``one of us (R. B. 
% G.) thanks $\ldots$''. Instead, try ``R. B. G. thanks$\ldots$''. Put sponsor 
% acknowledgments in the unnumbered footnote on the first page.

%%%%%%%%%%%%%%%%%%%% References %%%%%%%%%%%%%%%%%%%%
\bibliographystyle{IEEEtran}   % Use IEEE style for conferences
\bibliography{references}      % Link to your .bib file (without .bib extension)

\end{document}